\documentclass[letterpaper]{article} 
\usepackage{aaai24}  
\usepackage{times}  
\usepackage{helvet}  
\usepackage{courier}  
\usepackage[hyphens]{url}  
\usepackage{graphicx} 
\usepackage{natbib}  
\usepackage{caption} 
\usepackage{algorithm}
\usepackage{algorithmic}

\usepackage{times}
\usepackage{latexsym}
\usepackage{graphicx}
\usepackage{booktabs}
\usepackage{multicol}
\usepackage{multirow}
\usepackage{theorem}
\usepackage{makecell}
\usepackage{enumitem}

\usepackage{newfloat}
\usepackage{listings}
\DeclareCaptionStyle{ruled}{labelfont=normalfont,labelsep=colon,strut=off} 
\floatstyle{ruled}
\newfloat{listing}{tb}{lst}{}
\floatname{listing}{Listing}
\title{MedBench: A Large-Scale Chinese Benchmark\\for Evaluating Medical Large Language Models}
\author{
Yan Cai\textsuperscript{\rm 1},
Linlin Wang\textsuperscript{\rm 1,\rm 2}\thanks{Corresponding author.},
Ye Wang\textsuperscript{\rm 1},
Gerard de Melo\textsuperscript{\rm 3,\rm 4},
Ya Zhang\textsuperscript{\rm 2,\rm 5},
Yanfeng Wang\textsuperscript{\rm 2,\rm 5},
Liang He\textsuperscript{\rm 1}
}
\affiliations {
\textsuperscript{\rm 1}East China Normal University\\
\textsuperscript{\rm 2} Shanghai Artificial Intelligence Laboratory\\
\textsuperscript{\rm 3} Hasso Plattner Institute~~~~ \textsuperscript{\rm 4} University of Potsdam\\
\textsuperscript{\rm 5} Cooperative Medianet Innovation Center, Shanghai Jiao Tong University\\
\{51255901104,yewang\}@stu.ecnu.edu.cn, 
\{llwang,lhe\}@cs.ecnu.edu.cn,
gdm@demelo.org, 
\{ya\_zhang,wangyanfeng\}@sjtu.edu.cn}

\begin{document}

\maketitle

\begin{abstract}
The emergence of various medical large language models (LLMs) in the medical domain has highlighted the need for unified evaluation standards, as manual evaluation of LLMs proves to be time-consuming and labor-intensive. To address this issue,
we introduce MedBench, a comprehensive benchmark for the Chinese medical domain, comprising 40,041 questions sourced from authentic examination exercises and medical reports of diverse branches of medicine.
In particular, this benchmark is composed of four key components: the Chinese Medical Licensing Examination, the Resident Standardization Training Examination, the Doctor In-Charge Qualification Examination, and real-world clinic cases encompassing examinations, diagnoses, and treatments. 
MedBench replicates the educational progression and clinical practice experiences of doctors in Mainland China, thereby establishing itself as a credible benchmark for assessing the mastery of knowledge and reasoning abilities in medical language learning models. 
We perform extensive experiments and conduct an in-depth analysis from diverse perspectives, which culminate in the following findings: (1) Chinese medical LLMs underperform on
this benchmark, highlighting the need for significant advances in clinical knowledge and diagnostic precision. 
(2) Several general-domain LLMs surprisingly possess considerable medical knowledge.
These findings elucidate both the capabilities and limitations of LLMs within the context of MedBench, with the ultimate goal of aiding the medical research community.
 
\end{abstract}

\section{Introduction}
The advent of large language models (LLMs) has demonstrated substantial potential for diverse real-world applications, thanks to their remarkable language understanding capabilities. In the medical domain, a notable number of Chinese medical LLMs have successively emerged, including HuaTuo~\cite{wang2023huatuo}, ChatMed~\cite{zhu2023ChatMed}, BianQue~\cite{chen2023bianque1}, Sunsimiao~\cite{Sunsimiao}, and DoctorGLM~\cite{xiong2023doctorglm},  to better assist doctors in diverse tasks ranging from clinical diagnosis to disease prevention~\cite{Singhal_2023}. This underscores an urgent need for a standardized medical benchmark, capable of offering reliable and authoritative evaluations for such LLMs. Assessing the potential and inherent limitations of medical LLMs from diverse perspectives continues to present considerable challenges~\cite{Singhal_2023, chang2023survey}.

\begin{figure}[t]
    \centering
\includegraphics[width=0.95\columnwidth]{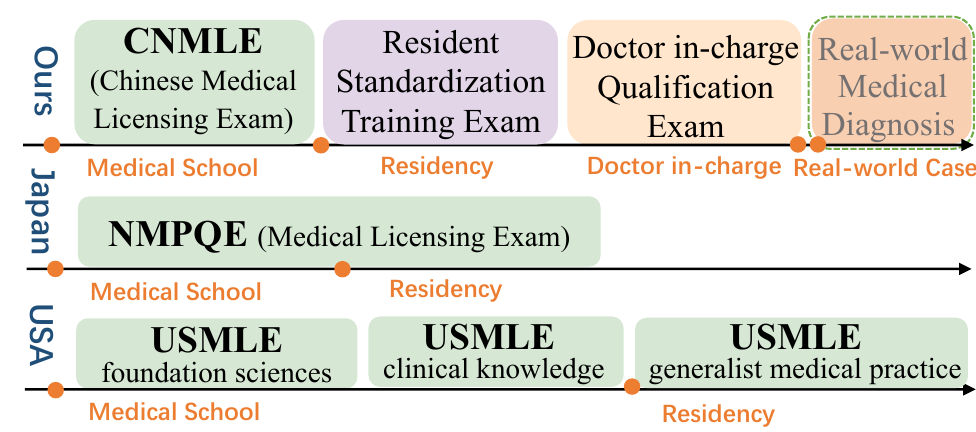}
    \caption{Comparison of procedures in different countries.}
    \label{fig:processes comparation}
\end{figure}

The primary cause of this issue lies in the pronounced discrepancy between the existing benchmarks and the practical realities of medicine, leading to an urgent need for advances in evaluation standards.
Widely used medical benchmarks such as MedQA~\cite{jin2021disease} and MedMCQA~\cite{pal2022medmcqa}
typically encompass publicly available medical question-answering datasets gathered from textbooks, research papers, and board exams. However, in light of recent research employing these open-access datasets for model training~\cite{han2023medalpaca}, a process potentially resulting in data contamination with regard to the evaluation~\cite{nori2023capabilities}, it is increasingly clear that the prevailing benchmarks exhibit considerable limitations. A recent study has made advances by introducing a human evaluation framework along with MultiMedQA~\cite{Singhal_2023}, a multi-faceted benchmark that combines existing QA datasets with a novel set of online medical questions. Nonetheless, given the disparities among typologically diverse languages, we are unable to rely on this benchmark for assessing Chinese large language models, in particular considering the unique clinical standards and procedures across different countries illustrated in Figure~\ref{fig:processes comparation}. 

\begin{figure*}
  \centering
\includegraphics[width=0.85\textwidth]{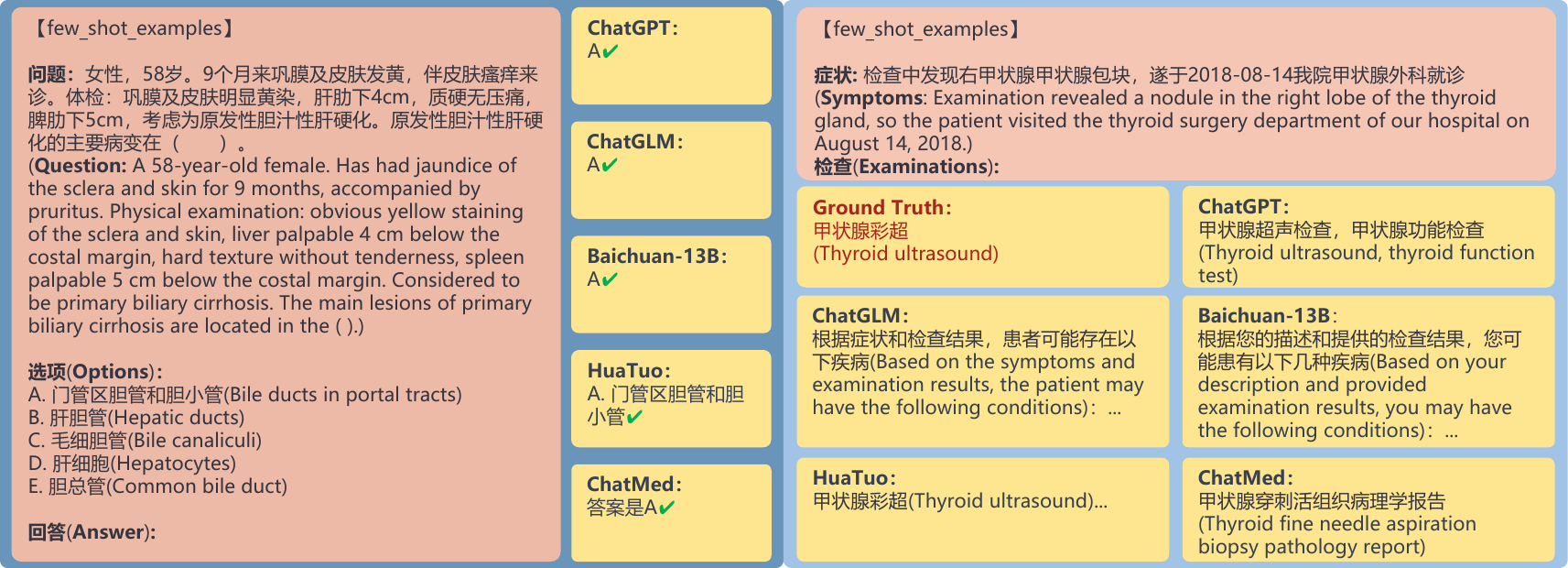}
  \caption{Examples of prompts and corresponding answers. The left side shows the prompt and responses of LLMs on an example question from an exam. The right side is an example of a real-world case.}
  \label{fig:coversation examples}
\end{figure*}

Hence, several Chinese benchmarks have been proposed, including a medical NLP-task oriented one\footnote{https://github.com/michael-wzhu/PromptCBLUE} and others based on the Chinese National Medical Licensing Examination (CNMLE), including MLEC-QA~\cite{li2021mlec} and CMExam~\cite{liu2023benchmarking}. While these benchmarks provide valuable insights, they are not exhaustive and may fall short in enabling us to comprehensively gauge the potential of LLMs with respect to all-round medical knowledge and their practical utility in real-world clinical diagnostic scenarios. In mainland China, a unique three-stage examination process is employed, key components of which, including the Resident Standardization Training and Doctor in-charge Qualification Exams, have largely been overlooked in prior work. Furthermore, the evaluation of real-world clinical practical skills has not been adequately incorporated in prior work.  

To address these gaps and align with the learning and growth trajectory of Chinese doctors, we introduce MedBench, a novel large-scale Chinese medical benchmark, which encompasses both authentic three-stage medical examinations and real-world clinical diagnosis cases. It surpasses prior benchmarks by being exclusively sourced from the latest validated exams and expert-annotated EHRs, thereby ensuring compliance with medical standards and practices. Furthermore, we conduct extensive experiments and offer detailed analyses to provide diverse perspectives for evaluating clinical knowledge recall and reasoning capabilities of LLMs across a range of branches of medicine. The main findings on this benchmark are as follows:
\begin{itemize}
    \item Chinese medical large language models underperform on this benchmark, necessitating substantial improvements in clinical knowledge and diagnostic accuracy, as well as refinement of their original in-context learning abilities.
    \item  Several large language models designed for general-domain tasks possess substantial medical knowledge, thereby exhibiting promising potential. 
    \item Human evaluation reveals that ChatGPT possesses rich knowledge for clinical practices, while current Chinese medical LLMs lack high-quality conversational abilities and sufficient medical knowledge.
    \item ChatGPT's unsatisfactory performance in case analysis, a question type that demands profound medical knowledge understanding and clinical reasoning abilities, reveals significant room for improvement in medicine.
\end{itemize}

\noindent In summary, this study presents a comprehensive benchmark that aligns with the practical realities of medicine in mainland China and provides profound empirical findings to reveal the medical capabilities and limitations of LLMs. Furthermore, we incorporate Item Response Theory~\cite{BIRNBAUM1969258} to further enhance our benchmark, with the ultimate goal of aiding the medical research community.

\begin{figure*}
    \centering
\includegraphics[width=0.85\textwidth]{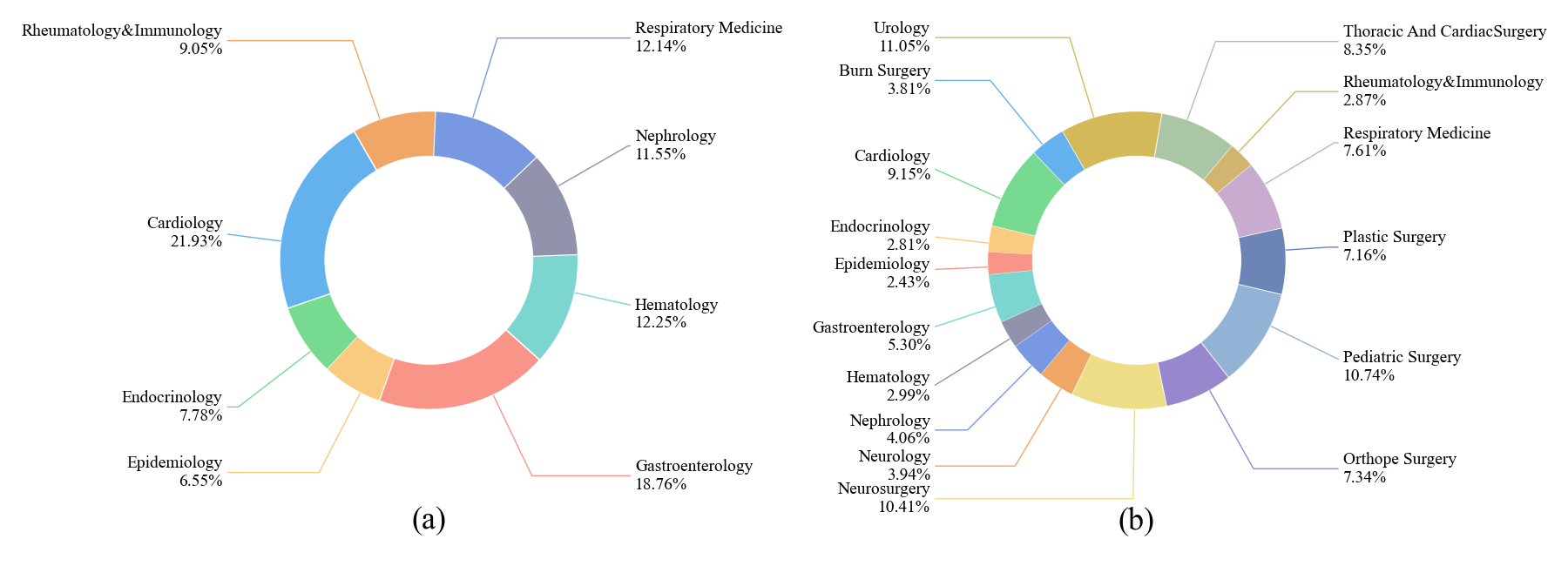}
    \caption{Branches of Medicine in MedBench. Part (a) represents the Resident Standardization Training Exam, and (b) reflects the Doctor in-charge Qualification Exam.}
    \label{fig:question sum}
\end{figure*}

\section{The Proposed Benchmark}
\paragraph{Overview}
For better evaluating medical foundation models, this paper proposes MedBench as a large benchmark with 40,041 exercises originating from both authentic medical examinations and real-world diagnostic and treatment cases. Specifically, we collect three-stage medical examinations that serve as a faithful reflection of the comprehensive process involved in obtaining medical licenses in mainland China, exemplifying essential medical knowledge. Moreover, we construct a number of real-world cases based on electronic health records that provide examination plans, diagnoses, and treatments based on patients' symptoms, which can reveal the medical knowledge utilization and reasoning capabilities of LLMs in the real world.

\paragraph{Construction and Statistics} Regarding the three-stage medical examinations, we collect representative exercises from the Chinese Medical Licensing Exam (CNMLE), Resident Standardization Training Exam, and Doctor in-charge Qualification Exam\footnote{http://www.nmec.org.cn/}
across many recent years, covering 27,248, 2,841, and 8,927 questions, respectively. The examination exercises, depicted 
on the left of 
Figure~\ref{fig:coversation examples}, are multiple-choice questions that can be categorized into three types:
\begin{itemize} \label{question type}
    \item A1/A2/B: Single statement questions with one correct answer out of five options.  
    \item A3/A4: A series of questions accompanied by a clinical case with one correct answer out of five options.
    \item Case Analysis: Given a clinical case, a series of questions are created with 6--12 options per question. Some questions may have more than one correct answer.
\end{itemize}
Figure~\ref{fig:question sum} depicts the classification of branches of medicine for some examinations in MedBench. Note that we use internal medicine and surgery as representative examples for the latter two stages, considering the vast number of subcategories within these fields. 

Furthermore, we collect over 2,000 real-world electronic health records, and employ experts to 
identify the symptoms, diagnoses, treatments, and examinations from the reports, resulting in
701 high-quality ones with an average length of 60--100 words per report (Figure~\ref{fig:coversation examples}). The annotators completed professional training and strictly following predefined annotation standards to ensure  
accuracy. Based on these annotated reports, a total of 1,025 question--answer pairs are eventually formed.

\paragraph{Characteristics} 
MedBench surpasses existing benchmarks in several aspects: (1) Authenticity and Novelty. It exclusively leverages 
expert-annotated EHRs and authentic up-to-date medical examinations to mitigate contamination. (2) Comprehensiveness and Multi-facetedness. It is meticulously designed to align with Chinese medical standards and practices by incorporating three-stage multi-disciplinary examinations and real-world clinical cases. (3) Practicality. Human evaluation on clinical real-world cases ensures congruence with practical realities of medicine, while difficulty-stratified divisions in MedBench enable rapid assessment.

\section{Evaluation}

\paragraph{Models and Evaluation Metrics}
To evaluate the medical capabilities, we conduct assessments using MedBench with several representative LLMs from both the general and medical domains, including ChatGPT, ChatGLM~\cite{zeng2022glm,du2022glm}, Baichuan-13B\footnote{https://github.com/baichuan-inc/Baichuan-13B.}, HuaTuo, and ChatMed. Furthermore, we evaluate other Chinese medical LLMs, such as BianQue, but we have observed that it lacks the ability to deliver accurate and reasonable responses to multiple-choice questions.

For the three-stage multiple-choice examinations, we employ accuracy as the evaluation metric. When dealing with real-world cases, we combine expert-level human evaluation with the additional automatic evaluation metrics BLEU~\cite{papineni2002bleu} and ROUGE~\cite{lin2004rouge}. 

\paragraph{Experimental Settings}
We conduct extensive experiments to evaluate the five-shot performance of LLMs, ensuring their capability to respond in a multiple-choice format. 
We leverage the API for ChatGPT\footnote{https://chat.openai.com.} and opt for local deployment to facilitate evaluations for other LLMs. 
Furthermore, we partition the MedBench dataset based on exams, medical subdiscipline, and question types, and perform independent testing on each subset to enable a comprehensive analysis.

\subsection{Main Results}
\subsubsection{Three-stage examination results.}
\begin{table*}[htpb]
\resizebox{\linewidth}{!}{
\begin{tabular}{lccccccccccc}
\Xhline{1.2pt}
\multirow{2}{*}{\textbf{LLM}} & \multicolumn{3}{|c|}{\textbf{CNMLE}} & \multicolumn{4}{c|}{\textbf{Resident Standardization Training }} & \multicolumn{4}{c}{\textbf{Doctor in-Charge Qualification}} \\ \cline{2-12}
\multicolumn{1}{l}{} & \multicolumn{1}{|c}{\textbf{Total}} & \textbf{A1/A2/B} & \multicolumn{1}{c|}{\textbf{A3/A4}} & \textbf{Total} & \textbf{A1/A2/B} & \textbf{A3/A4} & \multicolumn{1}{c|}{\textbf{Cases Analysis}} &  \textbf{Total} & \textbf{A1/A2/B} & \textbf{A3/A4} & \textbf{Cases Analysis} \\ \hline
\multicolumn{1}{l}{GPT-4} & \multicolumn{1}{|c}{\textbf{64.88}} & \textbf{63.08} & \multicolumn{1}{c|}{\textbf{69.03}} & \textbf{75.64} & \textbf{77.08} & \textbf{75.13} & \multicolumn{1}{c|}{\textbf{75.00}} & \textbf{68.45} & \textbf{71.91} & \textbf{68.24} & 62.80 \\
\multicolumn{1}{l}{ChatGPT} & \multicolumn{1}{|c}{49.57} & 49.40 & \multicolumn{1}{c|}{51.85} & 60.59 & 61.30 & 58.72 & \multicolumn{1}{c|}{62.96} & 58.75 & 58.04 & 59.73 & \textbf{65.52} \\
\multicolumn{1}{l}{ChatGLM} & \multicolumn{1}{|c}{27.39} & 27.32 & \multicolumn{1}{c|}{28.16} & 29.96 & 28.41 & 33.59 & \multicolumn{1}{c|}{29.63} & 27.52 & 26.06 & 31.43 & 28.97 \\
\multicolumn{1}{l}{Baichuan-13B} & \multicolumn{1}{|c}{30.47} & 30.54 & \multicolumn{1}{c|}{29.63} & 34.97 & 37.26 & 32.70 & \multicolumn{1}{c|}{22.65} & 29.56 & 31.31 & 31.65 & 17.89 \\
\multicolumn{1}{l}{HuaTuo} & \multicolumn{1}{|c}{22.31} & 22.38 & \multicolumn{1}{c|}{21.47} & 23.32 & 23.53 & 23.85 & \multicolumn{1}{c|}{13.58} & 21.93 & 22.14 & 21.85 & 18.62 \\
\multicolumn{1}{l}{ChatMed} & \multicolumn{1}{|c}{23.45} & 23.46 & \multicolumn{1}{c|}{23.33} & 24.33 & 23.03 & 26.54 & \multicolumn{1}{c|}{32.10} & 24.02 & 23.46 & 24.66 & 30.34 \\
\Xhline{1.2pt}
\end{tabular}}
\caption{Results on three-stage medical examinations of MedBench.}
\label{tab:overall results}
\end{table*}

In Table~\ref{tab:overall results}, we present a comprehensive analysis of the accuracy metrics for various LLMs across the three exams. A salient observation is that ChatGPT consistently surpasses other models, despite the latter being training on extensive Chinese corpora or premium medical datasets. Nevertheless, our observations indicate that ChatGPT's accuracy rate hovers around 50\% for CNMLE and approximately 60\% for other assessments, exposing substantial avenues for enhancement.

As depicted in Figure~\ref{MedEval_result}, ChatGPT exhibits subpar performance on questions pertaining to Traditional Chinese Medicine (TCM) and Chinese Western Medicine (CWM), with accuracy metrics oscillating between 40-45\%. Conversely, LLMs trained on a more expansive Chinese dataset demonstrate a narrower performance disparity on TCM and CWM questions relative to ChatGPT. This suggests that a contributing factor to ChatGPT's diminished efficacy on MedBench may be attributed to its limited exposure to Chinese data during its pretraining phase, consequently affecting its proficiency in Chinese medical knowledge.

It is imperative to underscore that Baichuan-13B and Hua\-Tuo encounter difficulties on case analysis questions, highlighting potential domains for refinement in their logical reasoning or multi-turn dialogue competencies.

We also perform comparative evaluations with a number of recently-developed models, including MedLLaMA~\cite{wu2023pmc}, Baize~\cite{xu2023baize}, and ChatDoctor~\cite{li2023chatdoctor}. However, we have not noticed any significant enhancements in performance on our benchmark compared to these models. For example, MedLLaMA achieves an accuracy of 24.5\% on the A1/A2/B type of questions, marginally outperforming ChatMed with a modest improvement of 1.5\%.

\paragraph{Real-world clinical case performance.}
\begin{figure*}
\begin{minipage}[b]{.31\linewidth}
    \centering
\includegraphics[width=0.75\columnwidth]{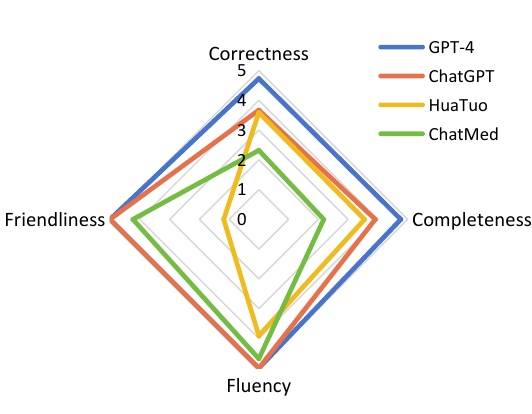}
    \caption{Human evaluation on clinical cases. }
    \label{fig:Human evaluation}
\end{minipage}
\begin{minipage}[b]{.03\linewidth}
~
\end{minipage}
\begin{minipage}[b]{.65\linewidth}

  \centering
  \resizebox{\linewidth}{!}{\begin{tabular}{lrrrrrrrrr}
    \toprule
    \multirow{2}{*}{LLM} & \multicolumn{3}{c}{BLEU-1} & \multicolumn{3}{c}{BLEU-4} & \multicolumn{3}{c}{ROUGE-L}\\
    \cmidrule{2-10}
    & E & T & D & E & T & D & E & T & D \\
    \cmidrule{1-10}
        GPT-4 & \textbf{16.54} & \textbf{11.25} & \textbf{12.17} & 3.09 & 1.66 & 2.36 & 0.95 & 0.82 & 2.66 \\
        ChatGPT & 11.50 & 8.37 & 7.29 & \textbf{3.26} & \textbf{1.82} & 1.85 & \textbf{15.39} & \textbf{11.30} & \textbf{10.04} \\
        ChatGLM & 0.77 & 2.12 & 1.90 & 0.11 & 0.22 & 0.22 & 1.79 & 3.62 & 3.29 \\
        Baichuan-13B & 11.07 & 5.29 & 10.14 & 1.50 & 0.52 & \textbf{2.59} & 0.71 & 0.0 & 1.11 \\
        HuaTuo & 1.56 & 3.77 & 2.32 & 0.23 & 0.42 & 0.15 & 2.43 & 0.52 & 1.06 \\
        ChatMed & 4.34 & 4.88 & 5.79 & 0.62 & 0.84 & 0.46 & 1.35 & 0.55 & 0.26 \\
    \bottomrule
  \end{tabular}}
  \captionof{table}{Evaluation on cases (E stands for Examinations, T stands for Treatments, and D stands for Diagnoses)}
  \label{tab:real-world cases results}

\end{minipage}
\end{figure*} 

Table~\ref{tab:real-world cases results} provides the results of the assessment on real-world cases, and Figure~\ref{fig:coversation examples} provides the prompt-response instances. We evaluated the outcomes using the BLEU and ROUGE F1-score metrics. It can be easily observed that ChatGPT and GPT-4~\cite{openai2023gpt4} exhibit superior performance, underscoring their remarkable aptitude in medical and conversational question answering. However, it is notable that even for ChatGPT and GPT-4, the top-performers across multiple trials, the scores remain relatively moderate. This could be attributed, in part, to the fact that metrics such as BLEU and ROUGE might not holistically capture result quality. Furthermore, considerable scope remains for enhancing these LLMs' capabilities in real-world clinical cases.

Beyond automated evaluations, the judgment of a postgraduate medical scholar was solicited to appraise the outputs delivered by the various LLMs on the real-world cases. The objective of this evaluation endeavor was to quantitatively assess the correctness, completeness, fluency, and friendliness of HuaTuo, ChatMed, ChatGPT, and GPT-4. Figure~\ref{fig:Human evaluation} provides the outcomes of this human evaluation. GPT-4 is found to consistently manifest superior performance across all delineated criteria, while ChatGPT marginally trails GPT-4, particularly on correctness and completeness. Conversely, HuaTuo's rating in friendliness is somewhat diminished, predominantly attributable to sporadic generation of incongruous content. However, it is important to note that HuaTuo demonstrates pronounced levels of correctness and completeness, evincing profound medical knowledge, though its articulation warrants improvement. ChatMed, while laudable for its fluency and friendliness, registers suboptimal scores in correctness and completeness. Such disparities might be indicative of a potential attenuation of medical expertise during the fine-tuning phase.

\section{Quantitative Analysis}
\subsubsection{Comparison on different question types.}

\begin{figure*}
    \centering
\includegraphics[width=\textwidth]{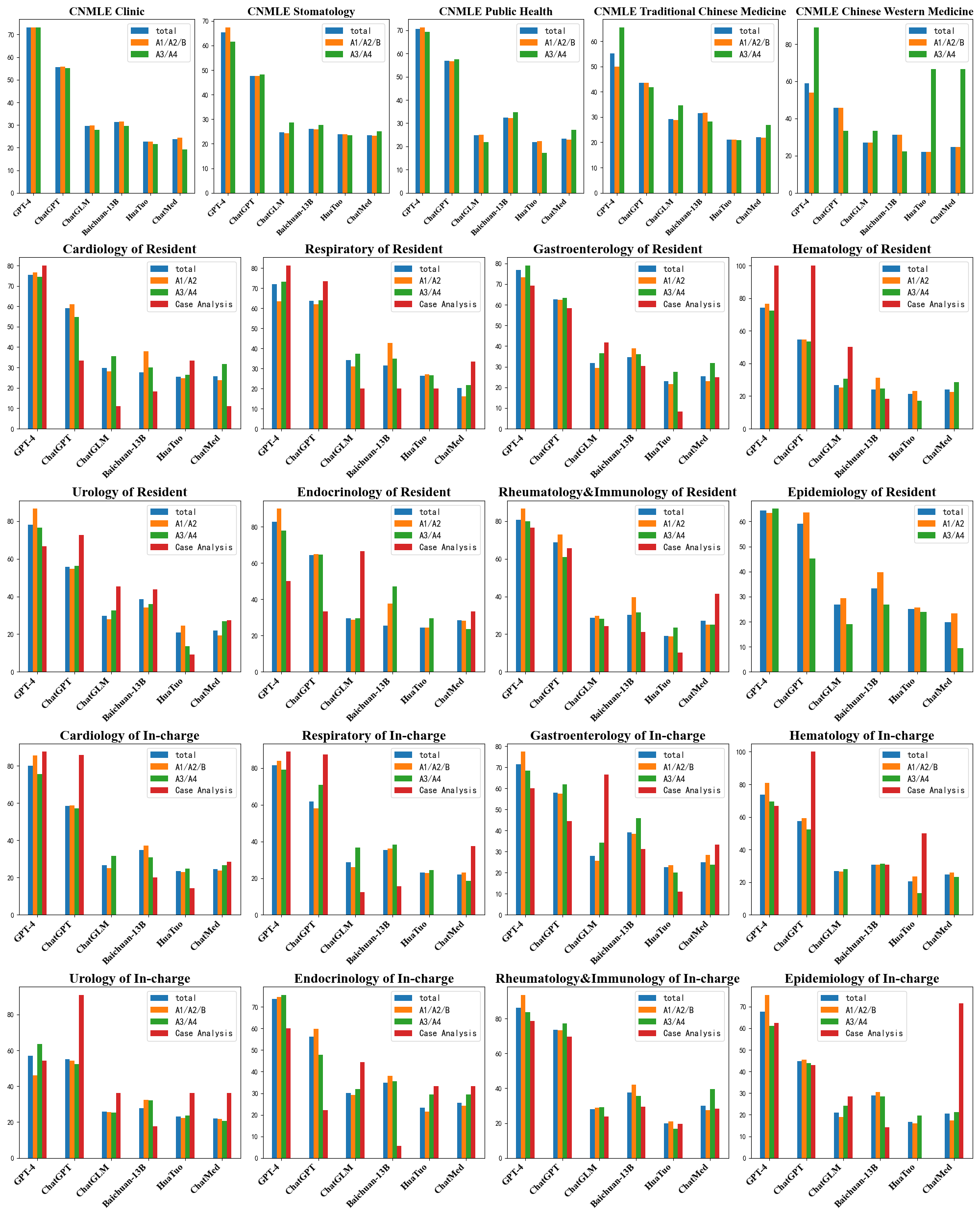}
    \caption{Results for three-stage medical examinations.}
    \label{MedEval_result}
\end{figure*}

Figure~\ref{MedEval_result} provides the detailed results on CNMLE, the Resident Standardization Training Exam (denoted as ``Resident'') and the Doctor in-charge Qualification Exam (``In-charge'').
Clearly, ChatGPT continues to exhibit a substantial performance advantage over the other LLMs. A3/A4 questions and case analysis questions, characterized by their multi-question structure, present a substantial challenge to the conversational abilities of the models. If a given LLM fares poorly on these question types, it indicates its inherent limitations in efficiently managing multi-turn dialogue or questions necessitating intricate reasoning steps.

\begin{figure*}
  \centering
  \includegraphics[width=0.85\textwidth]{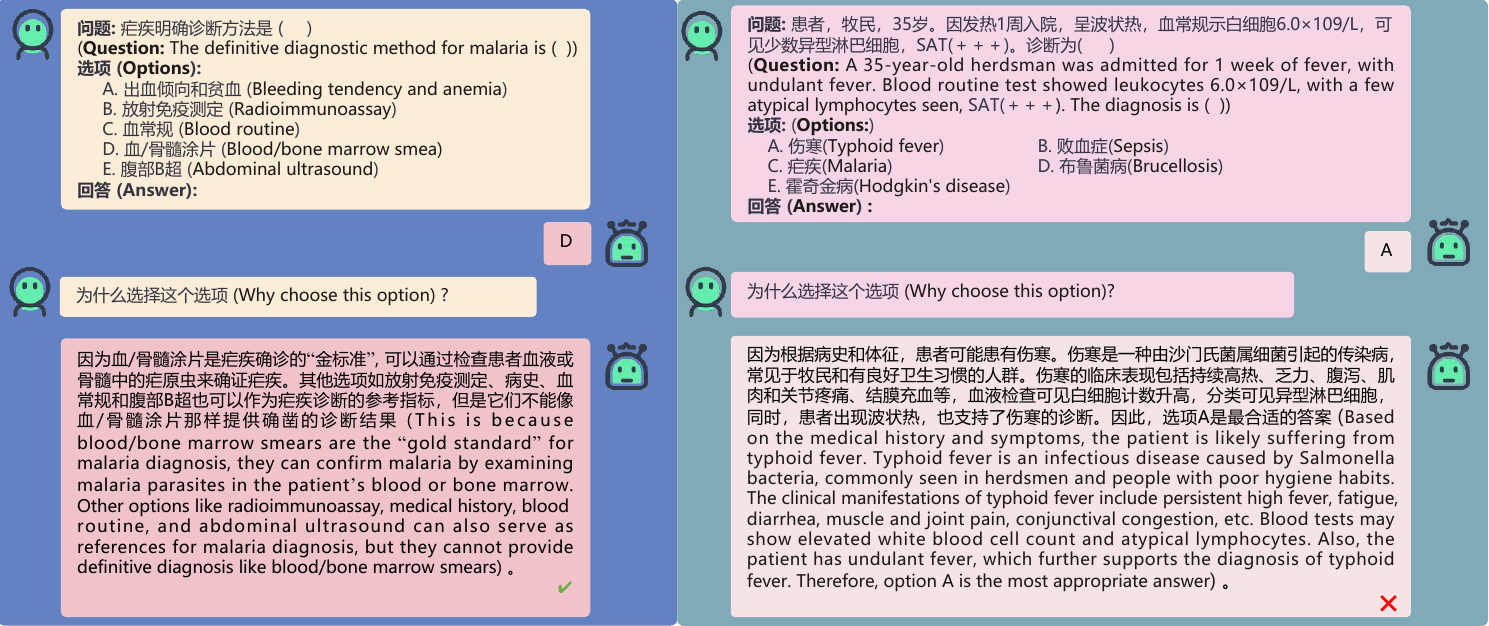}
  \caption{Examples of conversation demonstrating the reasoning capability of ChatGLM.}
  \label{fig:Interpretability examples}
\end{figure*}

\begin{figure}[h]
  \centering
  \includegraphics[width=0.85\columnwidth]{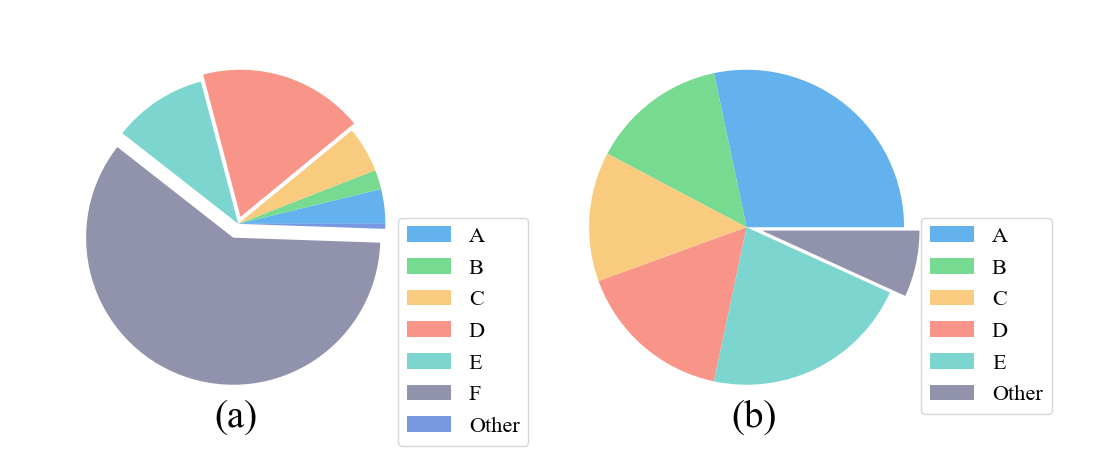}
  \caption{Distribution of choices made by Baichuan-13B under different prompts, where (a) and (b) depict the results using vanilla and Chain-of-Thought prompt, respectively.}
  \label{fig:Comparasion of different prompts}
\end{figure}

\begin{table}[H]
    \centering
    \begin{tabular}{ccc}
        \toprule
        \textbf{LLM} & \textbf{Vanilla prompt} & \textbf{Chain-of-Thought} \\
        \midrule
        ChatGLM-6B & 42.69 & 43.37 \\
        Baichuan-13B & 37.26 & 43.85 \\
        \bottomrule
    \end{tabular}
    \caption{A comparative assessment of Baichuan-13B and ChatGLM-6B using A1/A2 questions from the Resident Standardization Training Exam under different prompts.}
    \label{tab:result of different prompt}
\end{table}

\subsubsection{Chain-of-Thought}
In the experimental investigation, it is observed that prompts tailored to the LLM have the potential to enhance the reasoning capabilities of the LLM. Figure~\ref{fig:Comparasion of different prompts} illustrates the choice distribution of Baichuan-13B under both the vanilla prompt and Chain-of-Thought prompting. Notably, under the vanilla prompt, Baichuan-13B shows a strong inclination towards option F, although this option is not part of the valid choice set (A-E). In contrast, when using Chain-of-Thought prompting, Baichuan-13B primarily gravitates towards valid choices. Furthermore, Table~\ref{tab:result of different prompt} details the accuracy associated with both the vanilla prompt and the Chain-of-Thought one. The data suggests a significant improvement in the accuracy of Baichuan-13B when using Chain-of-Thought prompting. Parallel experiments are conducted on ChatGLM-6B, and the improvement in accuracy when using Chain-of-Thought prompting is negligible for ChatGLM-6B.

\subsubsection{Fewer questions, higher differentiation}
During the assessment process, some shortcomings became apparent: (1) Certain LLMs manifest suboptimal inference speed on GPUs or high computational cost when interfacing via an API. Given that MedBench comprises approximately 40,000 questions, performing inference on the entirety of this dataset consecutively takes considerable time for LLMs. (2) While categorization of questions based on the type provides a rudimentary gauge of difficulty, it is important to acknowledge that questions within the same category can exhibit disparate levels of difficulty. (3) Assigning particularly challenging questions to less capable LLMs is inadvisable, as it may culminate in uniformly diminished accuracy, making meaningful distinctions impossible.

\begin{figure}[t]
    \centering
    \includegraphics[width=0.95\columnwidth]{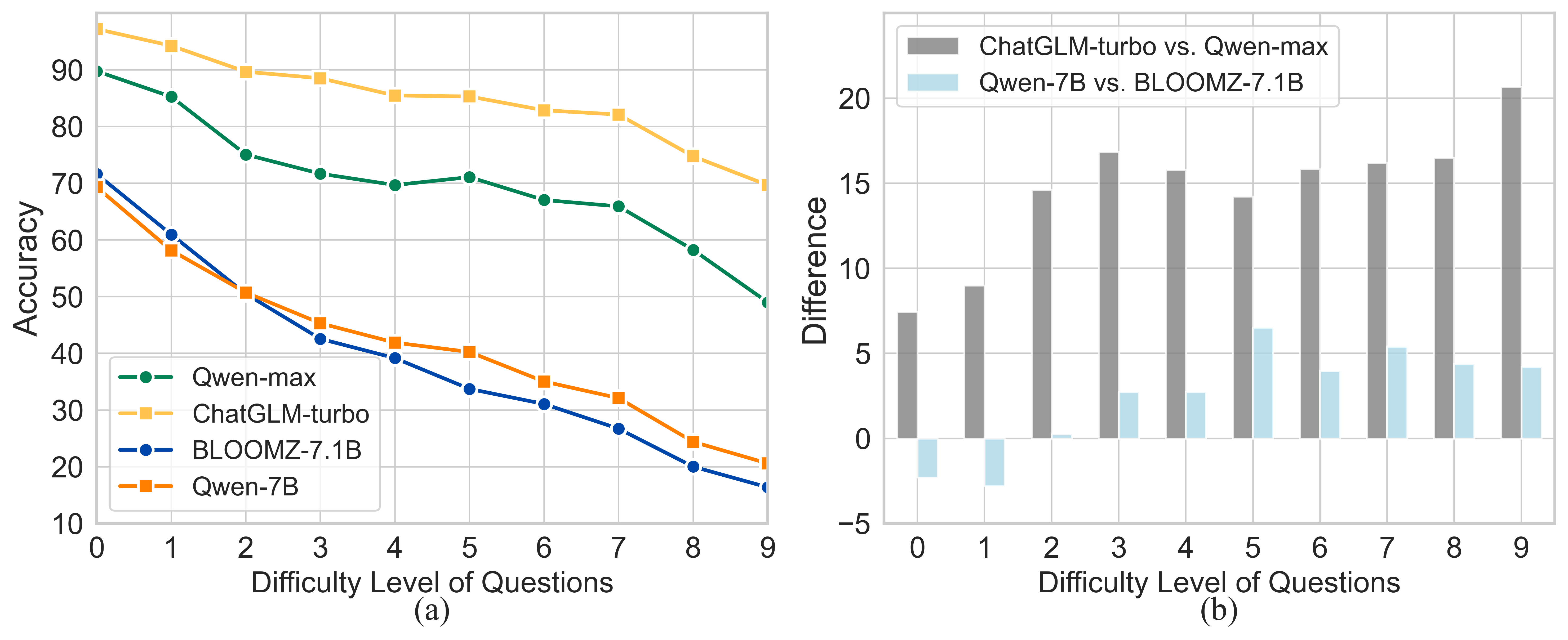}
    \caption{Evaluation of a few LLMs across varying levels of question difficulty, with Level 9 denoting the highest level of difficulty.}
    \label{fig:Difficulty Level of Questions}
\end{figure}

To ameliorate these challenges, we propose methodological strategies to classify questions of analogous types according to their inherent difficulty gradients. By adopting this paradigm, we can optimize the evaluation process, allowing LLMs to undertake inference on a curtailed set of questions. As a result, this allows for a more nuanced alignment between LLMs and questions of commensurate difficulty tiers, congruent with their individual proficiencies. Specifically, we introduce an advanced evaluative framework for LLMs drawing inspiration from Item Response Theory (IRT). Our approach integrates the three-parameter logistic model (IRT-3PL), given by:
\begin{equation} 
\label{eq:IRT}
P(X_{ij}=1|\theta_j) = c_i + (1-c_i)\frac{1}{1+e^{-a_i(\theta_j-b_i)}}
\end{equation}
Here, $\theta_j$ represents the proficiency of LLM $j$, and $P(X_{ij}=1|\theta_j)$ is the probability that an LLM $j$ with proficiency $\theta_j$ gives a correct response to question $i$. This equation highlights three crucial parameters for each question $i$: discrimination ($a_i$), difficulty ($b_i$), and the guessing factor ($c_i$). In our approach, we assume constant values for both $a_i$ and $c_i$, and then group items based on the shared difficulty metric $b_i$. For a specific difficulty subset, these three parameters remain unchanged. The equation captures the LLM's probability $\theta_j$, determined from the inference results when interacting with a specific subset. A total of 7,335 questions were divided into 10 difficulty levels. To validate question differentiation, we analyze them using BLOOMZ-7.1B~\cite{muennighoff2022crosslingual}, Qwen-7B\footnote{https://github.com/QwenLM/Qwen-7B.}, ChatGLM-turbo~\cite{zeng2022glm}, Qwen-max~\cite{qwen} across various difficulty levels. Part (a) of Figure~\ref{fig:Difficulty Level of Questions} displays the accuracy trends of these LLMs across these levels, highlighting a decrease in accuracy with increasing difficulty. 
Figure~\ref{fig:Difficulty Level of Questions}(b) shows the difference in accuracy between BLOOMZ-7.1B and Qwen-7B, as well as between Qwen-max and ChatGLM-turbo, for different levels of question difficulty. It is noticeable that the differences between the LLMs are significant when the question difficulty is appropriate.

\subsection{Reasoning Abilities}

In MedBench, there are several types of reasoning:

\begin{itemize}

\item \textbf{Multi-condition single-hop reasoning:}
A type of reasoning that requires the LLM to engage in single-hop reasoning based on a clinical scenario.

\item \textbf{Statement identification:}
A type of reasoning that requires the LLM to judge the correctness of multiple statements, which further tests the LLM's knowledge base.

\item \textbf{Multi-hop reasoning:}
A type of reasoning with multiple questions, where the questions are related, requiring the LLM to perform multi-hop reasoning.
\end{itemize}

\noindent To ascertain the extent to which LLMs make accurate decisions based on their grasp of pertinent knowledge, we extended our inquiry, 
which entailed requesting the LLMs to furnish justifications for their responses, as depicted in Figure~\ref{fig:Interpretability examples}. Our observations reveal that LLMs can substantiate their answers when they are accurate. Conversely, in cases where erroneous responses are given, the accompanying explanations often prove illogical, which provides compelling evidence that these LLMs either lack the requisite knowledge in the domain or are incapable of rationalizing towards the correct solutions.

\section{Related Work}
Traditionally, medical LLMs relied on classic medical QA benchmarks for evaluation. Some studies used USMLE for evaluation and achieved satisfactory results~\cite{kung2023performance, nori2023capabilities}, with zero-shot GPT-4 achieving average scores of 86.65\% and 86.7\% on the Self-Assessment and Sample Exam parts of USMLE, respectively. However, questions from USMLE have a very distinct Western medical perspective that is notably different from that of Chinese medicine. 
As two distinct medical systems, the latter includes traditional Chinese medicine, with obvious differences in concepts, diagnosis, and treatment methods. Therefore, the same symptoms may lead to different diagnoses and treatments. As a result, the USMLE cannot serve as an adequate benchmark for Chinese medical practice. MedQA, PubMedQA~\cite{jin1909dataset, lievin2022can}, and MLEC-QA are also useful benchmarks for medical QA, comprising a large number of high-quality questions, but they were proposed early on, potentially leading to data contamination. Although MedMCQA is a relatively new benchmark, its questions originate primarily from Indian medical institutions. As a result, analogous to USMLE, it lacks Chinese medical content and thus cannot serve to evaluate Chinese medical question answering sufficiently well. CMExam is a novel benchmark for medical QA in Chinese, comprising the latest exam questions from CNMLE, which has been manually annotated by medical experts with assistance from ChatGPT to ensure high quality. However, CMExam only comprises questions from CNMLE, while overlooking other major medical exams in mainland China. Moreover, as a predominantly multiple choice dataset with only a small portion of fill-in-the-blank questions, CMExam lacks real-world medical data and scenarios. This constrains its ability to fully assess LLMs' medical question answering capabilities. In fact, the optimal evaluation approach is to evaluate LLMs manually~\cite{xu2023medgpteval}. However, manual evaluation is a time-consuming and labor-intensive task, making it difficult to conduct on a large scale.

\section{Conclusion and Discussions}
In this paper, we present MedBench, an exhaustive benchmark specifically designed for the domain of Chinese medical question answering. Preliminary empirical analyses underscore the suboptimal performance of Chinese medical LLMs when subjected to this benchmark, highlighting the need for improved clinical acumen and diagnostic precision. Furthermore, the adeptness of these LLMs in contextual learning requires further refinement.

During our empirical investigations, we found that certain models manifest pronounced hallucinatory behavior. As our research progresses, it is imperative to ensure data veracity while delving into the systematic evaluation of such hallucinatory phenomena. Furthermore, the appraisal of the LLM's inferential competencies, as presented in this research, points to the need for further methodological refinements. Within the realm of clinical diagnostics, a diagnosis is typically predicated upon a plethora of corroborative evidence. In subsequent work, we plan to compile an enriched dataset, encompassing patients' antecedent medical records and comprehensive physical examination narratives, to strengthen the evaluative framework for medical LLMs. Simultaneously, our findings underscore the efficacy of psychometric methodologies in aiding the evaluation. Moving forward, we intend to enhance and further integrate such methodologies into our model assessment paradigm.

\bibliography{aaai24} 

\end{document}